\documentclass[11pt,a4paper]{article}
\usepackage[margin=25mm]{geometry}
\usepackage[T1]{fontenc}
\usepackage[utf8]{inputenc}
\usepackage{mathptmx,microtype,amsmath,amssymb,booktabs,graphicx}
\usepackage{caption,placeins,xcolor,longtable}
\usepackage[numbers,sort&compress]{natbib}
\usepackage[colorlinks=true,linkcolor=blue!45!black,citecolor=blue!45!black,urlcolor=blue!45!black]{hyperref}
\hypersetup{pdftitle={AnchorFlow-VIO: Fixed-Template Monocular Visual-Inertial Odometry on LaMAria},pdfauthor={Sudhir Pratap Yadav}}
\title{AnchorFlow-VIO: Fixed-Template Monocular Visual-Inertial Odometry on LaMAria}
\author{Sudhir Pratap Yadav}
\date{}

\begin{document}
\maketitle
\begin{abstract}
We present AnchorFlow-VIO, an open-source Basalt-derived monocular visual-inertial
odometry system, and evaluate a uniform configuration on LaMAria. The system uses
one grayscale camera and one IMU, combining
factory IMU correction, a fixed timing adjustment, windowed initialization and
fixed-template patch tracking. Server evaluation of all 63 test recordings yields
control-point scores of 47.2, 37.1, 24.8, 35.1 and 15.6 for Short, Medium, Long,
Low light and Moving platform. Test pose recalls at 5 m are 84.6\%, 69.1\%,
53.9\% and 62.9\% for the four categories reporting dense recall. On ten training
recordings, mean control-point score is 41.38 and all 265,694 image timestamps
receive poses, including 204 filled poses. One training platform recording
diverges and scores zero.
A paired four-record comparison favors fixed over rolling templates on three
recordings, while reversing that ranking on the fourth. Increasing feature
density also yields mixed results. Ten fresh runs at prospectively fixed 30 s
and 120 s offsets expose substantial start sensitivity: a low-light recording
scores 6.14 at the earlier start and 42.00 at the later start. These results
characterize the limits of the frozen system and show why complete output,
feature counts and selected successful runs are insufficient evidence of robust
localization. Source code, frozen configuration and reproducible submission
tools are publicly available; training ablations are reported separately from
the submitted test evaluation.
\end{abstract}

\section{Introduction}
Head-mounted visual-inertial estimation must accommodate rapid viewpoint
changes, poor illumination and scenes in which visible objects move independently
of the observer. LaMAria provides a setting for studying these conditions at city
scale, including low-light and moving-platform challenges
\cite{krishnan2025lamaria}. The present report focuses on a restricted sensor
configuration: the left grayscale camera and a single IMU from the ordinary ASL
recordings. A second camera is neither supplied to the estimator nor used to
generate its initial state or landmarks.

The practical question is how well an existing visual-inertial estimator transfers
to this setting under a uniform configuration. A result on an easy recording
does not determine performance on long walks, dark scenes or platform motion.
Likewise, successful process termination and a pose at every image timestamp do
not establish accurate tracking. Output can remain finite and complete while
the estimated trajectory drifts or diverges.

We study AnchorFlow-VIO, a frozen Basalt-derived monocular visual-inertial
odometry (VIO) configuration. Its main estimator architecture is inherited from prior work
\cite{usenko2019basalt,demmel2021sqrt}. Our contribution is an empirical system
characterization: a sensor-isolated implementation, paired template and feature
density comparisons, complete evaluation on ten challenge training recordings,
and fresh-start tests that retain the unfavorable outcomes. A public source
release and server evaluation on all 63 test recordings extend this characterization
beyond the ten training recordings. The benchmark lists the method as
\emph{AnchorFlow VIO} \cite{anchorflow_method}. We do not introduce
a new estimator formulation or claim that fixed-template tracking is a new
general principle.

Four questions organize the study. First, how does the frozen policy perform
across challenge categories? Second, do template retention and increased feature
density improve the same recordings? Third, how much do results change when
initialization occurs later in an otherwise identical recording? The last
comparison uses matched suffixes of the original trajectories so that different
temporal coverage is not mistaken for a method effect. Fourth, how does the
unchanged configuration perform in the benchmark's server-side test evaluation?

\section{Related Work and Scope}
Basalt combines visual-inertial odometry with a mapping formulation based on
nonlinear factor recovery \cite{usenko2019basalt}. Square root marginalization
provides a sliding-window formulation that avoids forming normal equations
during marginalization \cite{demmel2021sqrt}. We use the Basalt-derived VIO
estimator and its square root configuration; the evaluated candidate does not
include the original global mapping pipeline. IMU preintegration is a standard
way to summarize measurements between visual states; Forster et al. develop
its manifold formulation \cite{forster2015preintegration}. These estimator ideas
are inherited components rather than contributions of this report.

ORB-SLAM3 supports visual, visual-inertial and multiple-map estimation, including
monocular operation \cite{campos2021orbslam3}. Our frozen candidate has no
validated loop closure or global relocalization extension. We therefore refer
to it as odometry, even though its output format is intended for a SLAM
benchmark. We do not compare its training scores with other methods' hidden-test
scores as evidence of relative performance.

LaMAria introduces egocentric recordings with survey-based control-point
evaluation \cite{krishnan2025lamaria}. This report supplies a focused analysis
of one implementation on ten challenge training recordings and 63 test recordings.
It does not replace
the benchmark's broader comparison, reproduce every published monocular
baseline, or establish novelty over all prior template-tracking methods.

\section{System}
\subsection{Sensor isolation and preprocessing}
The input view contains only \texttt{cam0} and \texttt{imu0}. The former is the
left grayscale camera in the benchmark's undistorted pinhole representation;
the latter is the right 1 kHz IMU. Camera choice is fixed across recordings.
The native reader returns exactly one image per timestamp, the calibration
contains one camera, and dumped observations are checked for camera ID zero.
The pipeline regenerates states and tracks from these inputs. It does not reuse
stereo trajectories, depths, landmarks or initialization states.

For each gyroscope or accelerometer sample, the factory affine correction is
\begin{equation}
 \mathbf{u}_{\mathrm{corr}}=M^{-1}(\mathbf{u}_{\mathrm{raw}}-\mathbf{b}_{\mathrm{factory}}).
 \label{eq:calibration}
\end{equation}
Here $M$ and $\mathbf{b}_{\mathrm{factory}}$ are device-specific factory
calibration parameters. The preparer subtracts $4.2\times10^6$ ns from IMU
timestamps. This fixed adjustment and the noise settings are transferred from
the parent development; they are not re-estimated for individual mono runs.
Relative to the calibration conversion's base noise values, accelerometer and
gyroscope noise multipliers are 20 and 2, and their bias random-walk multipliers
are 1 and 0.1. These are empirical configuration choices, not a claim that the
sensor's physical noise has those magnitudes. An unknown factory-calibration
identifier causes preparation to fail rather than silently applying an identity
correction.

\subsection{Fixed-template tracking}
The frontend uses pyramidal patch alignment. In the retained policy, each track
stores its birth-template patches and aligns that reference to subsequent
images, with the current transform serving as an initial estimate. The
implementation accounts for the patch's birth linear transform and performs
alignment over the configured pyramid. The rolling-template control instead
uses the preceding image as its temporal reference. Their paired comparison
changes only the template-policy switch.

Retaining a reference template can avoid repeated updates to the reference
appearance, but can also make matching harder after substantial appearance or
viewpoint change. These are possible explanations for a tradeoff, not mechanisms
identified by the present experiments. We assess the resulting trajectories
rather than infer accuracy from the number of surviving features.

The detection grid size is 50 pixels in the frozen policy. The dense-feature
variant changes it to 35 pixels, leaving the remaining policy unchanged. The
configuration uses patch pattern 51, a pyramid-level parameter of 4, and at most
5 optical-flow iterations. Both-camera detection flags inherited from the
parent configuration cannot create a second stream in the isolated reader.
Stereo initialization is disabled.

\subsection{Initialization and sliding-window estimation}
Initialization collects a nominal one-second window. Accelerometer samples are
rotated into a common frame using integrated gyroscope rotation before averaging.
The selected initialization mode uses the resulting gravity direction in a
linear visual-inertial velocity solve. The implementation can move the visual
reference to a frame with surviving features. If the solve is rejected, it falls
back to the averaged-acceleration attitude and zero velocity. We retain this
policy without choosing an initializer separately for each recording.

The estimator jointly maintains poses, velocities, inertial biases and local
landmark information within the inherited square root VIO architecture. The
native configuration uses \texttt{ABS\_QR}, square root marginalization,
\texttt{vio\_max\_states=3}, \texttt{vio\_max\_kfs=10}, and at most seven VIO
iterations. Observation standard deviation and Huber threshold are configured
as 0.5 pixels and 1.0, respectively, with an outlier threshold of 3 pixels.
These settings specify the evaluated system; none is independently validated
as an optimal setting for monocular egocentric motion.

\subsection{Trajectory export and coverage}
Output is the world-from-IMU transform, with ordered integer nanosecond
timestamps, translation and quaternion components in $xyzw$ order. The converter
fills the initial interval with the first available native pose and later missing
timestamps with the preceding pose. The first rule uses future availability, so
the complete exported pipeline is not wholly causal. Filled poses satisfy the
format contract but add no visual evidence. We report coverage separately from
accuracy and record initial and later fills.

\section{Experimental Protocol}
\subsection{Training recordings and selection}
The full-record evaluation includes two training recordings in each of five
categories: Short, Medium, Long, Low light and Moving platform. We abbreviate
\texttt{sequence\_1\_19} as Short19, with corresponding labels for the other
records in Table~\ref{tab:full}. The four-record exploratory screen comprises
Short19, Medium12, Long17 and Lowlight10. The remaining six records are the
prospective expansion of the feature-density comparison; they are not an
untouched holdout from the wider parent research program.

The original fixed-template policy is retained uniformly. No per-record winner
is substituted into the exported candidate. Variants using rolling templates,
denser features, global refinement, rotating-world models and recovery remain separate
experiments. Ground truth is used after trajectory creation for evaluation,
not supplied to the native inference dataset. Training results support
configuration development; the submitted test trajectories use the same frozen policy.

\subsection{Test submission and public evaluation}
The frozen method was run on all 63 main-test recordings: 18 Short, 10 Medium,
16 Long, 9 Low light and 10 Moving platform. Individual trajectories and the
combined archive passed the official parser and full original image-timestamp
coverage checks. The archive contains 1,984,414 poses. Test diagnostics exposed
large steps on several platform recordings; the original outputs were retained
without parameter changes or trajectory repair based on those observations.
An interrupted execution was preserved and rerun with the identical frozen
method; this was execution recovery, not score-based run selection.

The owner uploaded the combined archive and supplied the server's per-sequence
result display. Public method ID 45 confirms the upload on 19 October 2026
\cite{anchorflow_method}. We independently checked the live category leaderboard
on the same date \cite{lamaria_leaderboard}. Headline test values below reproduce
that display; Appendix~\ref{sec:test_appendix} preserves all 63 per-record values
at the displayed one-decimal precision. Test accuracy is computed by the server,
not locally from unavailable test reference trajectories. We keep the submitted
test evaluation separate from training ablations and fresh-start studies.

\subsection{Metrics and alignment}
For local training evaluation, we use the official implementation, pinned at commit
\texttt{aa811b3}. Control-point (CP) score is the official normalized piecewise
linear score of horizontal control-point reconstruction errors. In the checked
implementation, its error/score knots are
\begin{equation}
 (0,20),\ (0.05,20),\ (0.20,18),\ (0.50,15),\ (1,12),\ (2,8),\ (5,4),\ (10,0),
\end{equation}
where errors are in meters. Per-point scores are divided by 20 and averaged
to obtain a score on a 0--100 scale. Failed point triangulations contribute
zero through a large-error substitution. A failed overall alignment remains
an unavailable score and is not silently converted into an observed zero.

Pose recall at 5 m uses horizontal position error after the control-point
similarity alignment, divided by the number of reference poses. Absolute
trajectory error (ATE) is separately computed against pseudo ground truth with
the official MPS evaluator and similarity alignment. We report its RMSE and the
fitted Sim(3) scale. Thus ATE and pose recall do not use interchangeable
alignment procedures. A scale different from one describes the fitted
correction to the estimated trajectory and is not a sensor-calibration estimate.
The training platform records have no dense pseudo ground truth, so recall, ATE
and scale are unavailable. Training category values are unweighted means over
their two recordings; the training CP mean weights all ten recordings equally.
Server category scores and recall are reported separately at official display
precision. Test per-record ATE was not returned in the supplied result display
and is not inferred from CP or recall. Platform test recall is absent from the
summary; an isolated zero in the per-record display is retained only in the appendix.

\subsection{Fresh starts and execution conditions}
The controlled-start plan was frozen at commit \texttt{9acc71b}. Five recordings
cover the five categories, each with fresh starts at 600 and 2,400 frames,
nominally 30 s and 120 s at 20 Hz. Each run starts a new process on the complete
remaining raw mono input; no state or landmark is carried from the original run.
Its comparator is the original image-zero trajectory restricted to exactly
the same timestamps. Official recall retains the full-reference denominator,
including the missing-prefix penalty. Consequently, suffix comparisons are
meaningful, while direct comparison with full-record recall would confound
temporal coverage.

Runs use an Intel Core i7-10700 CPU and two estimator threads. The controlled
starts use CPUs 14 and 15, a 4 GiB service memory cap and Nice 15, with a single
native worker. Affinity is enforced because the requested CPU quota was not
effective on this host. The initial exploratory timing predates a uniform
affinity policy. We report standardized controlled-start timing separately and
do not use early wall times to establish a template-speed advantage.

These are recorded runs of the deterministic-mode configuration, not independent
random-seed repetitions. Start offsets vary the input interval and initialization;
they are not repeated samples from an independent population. We therefore
report descriptive paired results without confidence intervals or significance
tests over correlated frames.

\section{Results}
\subsection{Server-evaluated test performance}
\begin{figure}[t]
\centering\includegraphics[width=\linewidth]{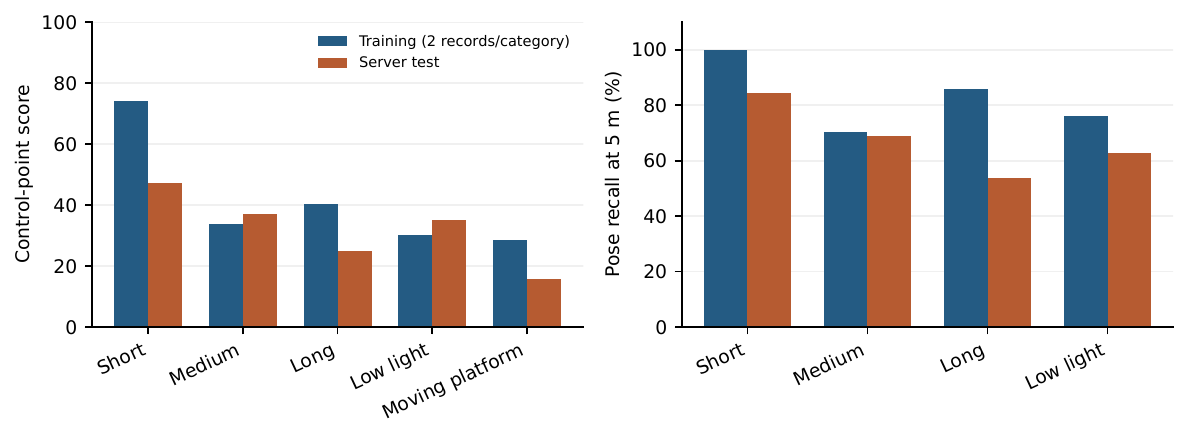}
\caption{Training and test category scores and pose recall. Training means use
two recordings each; test values are the official category display for the
18/10/16/9/10-record submission. Platform recall is unavailable in the summary.}
\label{fig:training_test}
\end{figure}

\begin{table}[t]
\centering\small\begin{tabular}{lrrrrr}\toprule & Test & \multicolumn{2}{c}{CP score} & \multicolumn{2}{c}{R@5m (\%)} \\ Category & records & Training & Test & Training & Test \\\midrule
Short & 18 & 74.05 & 47.2 & 100.00 & 84.6 \\
Medium & 10 & 33.68 & 37.1 & 70.48 & 69.1 \\
Long & 16 & 40.37 & 24.8 & 85.99 & 53.9 \\
Low light & 9 & 30.16 & 35.1 & 76.18 & 62.9 \\
Moving platform & 10 & 28.65 & 15.6 & \textemdash & \textemdash \\
\bottomrule
\end{tabular}

\caption{Training and server-evaluated test results for the frozen method.
Training has two recordings per category; test counts differ as shown.
Different recordings prevent interpreting the differences as paired effects.}
\label{tab:test_summary}
\end{table}
AnchorFlow-VIO obtains test CP scores of 47.2, 37.1, 24.8, 35.1 and 15.6 across
Short, Medium, Long, Low light and Moving platform. Test recalls are 84.6\%,
69.1\%, 53.9\% and 62.9\% where reported. Compared with the training category
means, test CP is lower on Short, Long and Moving platform but higher on Medium
and Low light (Table~\ref{tab:test_summary}, Figure~\ref{fig:training_test}).
These are different recording sets and sample sizes, so this comparison
describes performance variation rather than a paired generalization estimate.

\begin{table}[t]
\centering\footnotesize\setlength{\tabcolsep}{4pt}
\begin{tabular}{lrrrrr}\toprule Method & Short & Medium & Long & Low light & Platform \\\midrule
Mighty Camera & 57.1 / 88.3 & 38.8 / 74.8 & 18.5 / 39.3 & 36.9 / 64.2 & 15.8 \\
\textbf{AnchorFlow-VIO} & 47.2 / 84.6 & 37.1 / 69.1 & 24.8 / 53.9 & 35.1 / 62.9 & 15.6 \\
FPV-SLAM & 47.0 / 81.0 & 28.4 / 67.0 & 25.7 / 54.6 & 45.0 / 80.7 & 22.5 \\
ASTRM\_FLOWSLAM\_v1 & 46.9 / 76.2 & 30.1 / 56.7 & 26.0 / 58.1 & 34.3 / 59.3 & 23.5 \\
AnonSLAM & 43.3 / 91.9 & 33.6 / 78.9 & 30.0 / 76.9 & 41.7 / 88.3 & 19.1 \\
(r) OpenVINS+Maplab & 19.7 / 43.8 & 12.7 / 29.0 & 5.5 / 10.2 & 6.9 / 15.0 & 2.0 \\
(r) ORB-SLAM3 & 23.0 / 61.2 & 10.9 / 26.0 & 11.2 / 28.7 & 3.1 / 9.0 & 2.0 \\
(r) OpenVINS & 16.0 / 40.3 & 10.6 / 24.0 & 5.5 / 13.7 & 5.5 / 13.0 & 1.5 \\
\bottomrule
\end{tabular}

\caption{Selected monocular + IMU entries from the official leaderboard on
19 October 2026. Cells show CP / R@5m (\%); Platform shows CP only. Parenthetical
``r'' denotes organizer-reproduced runs. Values are server results, not reruns
by the author, and no speed comparison is implied.}
\label{tab:leaderboard}
\end{table}
On the dated leaderboard, AnchorFlow-VIO has the second-highest Short CP among
monocular + IMU entries, behind Mighty Camera. Its Short score exceeds the
monocular FPV-SLAM entry by 0.2 CP, while FPV-SLAM has higher Long, Low-light
and Platform scores (Table~\ref{tab:leaderboard}). Ranking therefore depends
on the challenge and metric; we do not claim a single overall rank or an
exhaustively verified open-source-method rank. The public implementation makes
the submitted configuration available for further comparison.

All 63 outputs retain their original coverage, but several test estimates have
severe discontinuities. Maximum consecutive steps are approximately 181.23 m,
125.07 m and 108.61 m on \texttt{sequence\_5\_4}, \texttt{sequence\_5\_5} and
\texttt{sequence\_5\_6}; their server CP scores are 0.0, 0.0 and 8.4.
These cases show that full-format coverage does not establish reliable tracking
on the submitted test data either. Per-sequence test scores are descriptive
outcomes of the frozen submission and are not used to select a revised policy.

\subsection{Complete training-record performance}
\begin{table}[t]
\centering\small
\begin{tabular}{lrrrrr}\toprule Record & CP & R@5m (\%) & ATE (m) & Scale & Poses \\\midrule
Short19 & 79.02 & 100.00 & 0.59 & 0.9854 & 18,352 \\
Short20 & 69.08 & 100.00 & 1.03 & 0.9789 & 20,262 \\
Medium11 & 49.53 & 94.78 & 2.16 & 1.0051 & 23,748 \\
Medium12 & 17.83 & 46.18 & 7.01 & 0.9508 & 33,513 \\
Long17 & 48.90 & 92.97 & 2.44 & 0.9999 & 35,842 \\
Long18 & 31.85 & 79.00 & 4.01 & 0.9850 & 33,878 \\
Lowlight10 & 23.18 & 64.64 & 5.36 & 0.9997 & 28,410 \\
Lowlight11 & 37.13 & 87.73 & 3.79 & 1.0034 & 22,541 \\
Platform11 & 57.30 & \textemdash & \textemdash & \textemdash & 22,483 \\
Platform12 & 0.00 & \textemdash & \textemdash & \textemdash & 26,665 \\
\bottomrule
\end{tabular}

\caption{Frozen monocular policy on all ten challenge training recordings.
R@5m is horizontal pose recall. ATE and scale use separate Sim(3) trajectory
alignment. Dashes indicate unavailable dense reference metrics.}
\label{tab:full}
\end{table}
The frozen policy obtains mean CP 41.38302 over ten complete recordings.
Category CP scores are 74.05, 33.68, 40.37, 30.16 and 28.65 for Short, Medium,
Long, Low light and Moving platform. Corresponding mean recalls are 100.00\%,
70.48\%, 85.99\% and 76.18\% for the four categories with dense references.
Table~\ref{tab:full} shows the variation hidden by these averages: Short19 scores
79.02, whereas Medium12 scores 17.83. Platform11 reaches 57.30 but Platform12
diverges and scores zero.

\begin{figure}[t]
\centering\includegraphics[width=\linewidth]{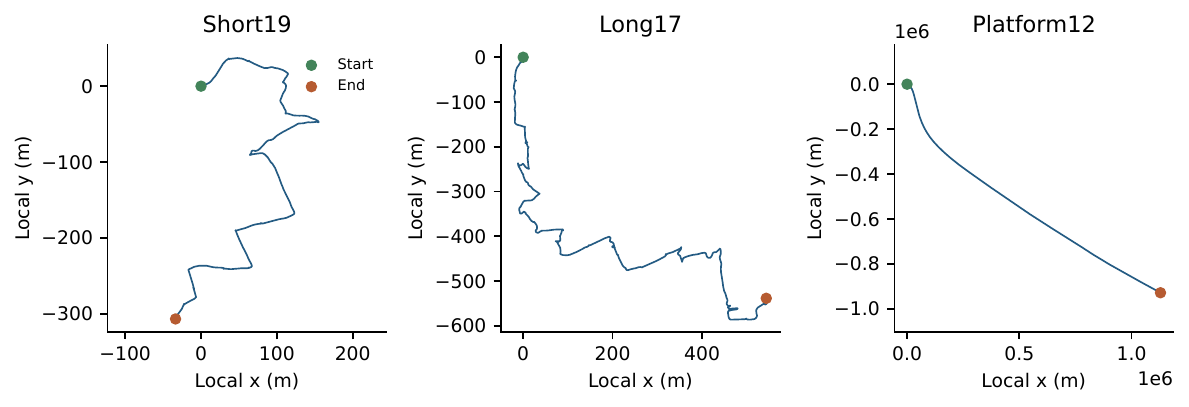}
\caption{Estimated horizontal trajectories after subtracting each initial
position. No ground truth, yaw alignment or scale fitting is applied. Axes are
independent between panels. Platform12's extreme extent exposes the failed
estimate rather than a plausible traveled path. Green and orange markers show
the first and last positions.}
\label{fig:trajectories}
\end{figure}

All 265,694 requested timestamps are represented, including 200 initial fills
and four later fills. Platform12 preserves this coverage despite a maximum
native step of 139.25 m. This is a numerical trajectory failure rather than
successful localization. Figure~\ref{fig:trajectories} shows selected estimated
paths, including the failed recording. These are unaligned paths in separate
local coordinate systems; their shapes do not measure accuracy.

\subsection{Template choice and scale}
\begin{table}[t]
\centering\small\begin{tabular}{lrrrr}\toprule & \multicolumn{2}{c}{CP} & \multicolumn{2}{c}{R@5m (\%)} \\ Record & Fixed & Rolling & Fixed & Rolling \\\midrule
Short19 & 79.02 & 54.46 & 100.00 & 97.23 \\
Medium12 & 17.83 & 33.26 & 46.18 & 78.64 \\
Long17 & 48.90 & 17.13 & 92.97 & 35.68 \\
Lowlight10 & 23.18 & 20.78 & 64.64 & 48.92 \\
\bottomrule
\end{tabular}

\caption{Paired fixed versus rolling templates on four exploratory recordings.
All other policy settings are unchanged.}
\label{tab:templates}
\end{table}
Fixed templates improve CP over rolling templates by 24.57 on Short19, 31.77 on
Long17 and 2.41 on Lowlight10, but lose 15.43 on Medium12
(Table~\ref{tab:templates}). The corresponding recall changes also reverse
direction across records. Fixed templates therefore cannot be characterized as
uniformly superior from this screen. The retained configuration is an existing
uniform policy, not a composite of the best four outcomes.

\begin{figure}[t]
\centering\includegraphics[width=\linewidth]{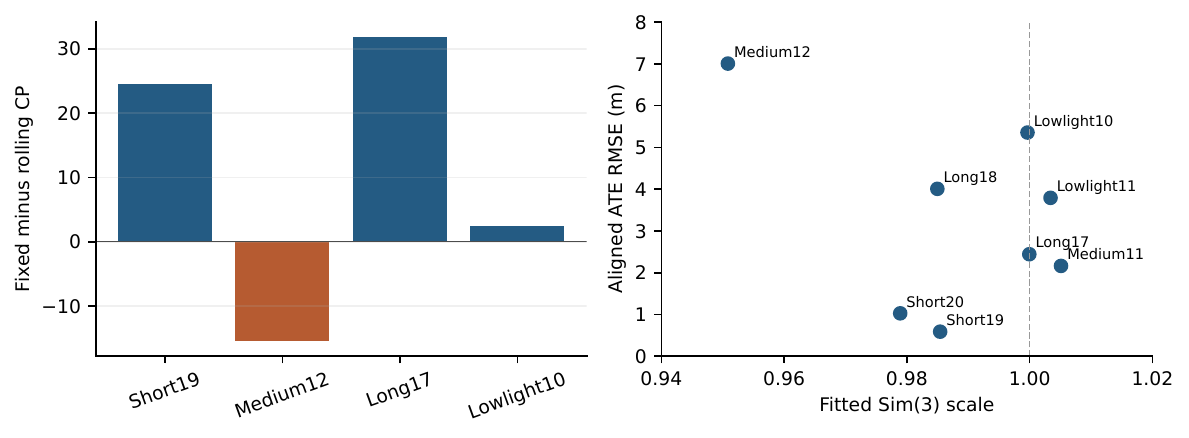}
\caption{Left: fixed-minus-rolling CP difference on the exploratory screen.
Right: fitted similarity scale and aligned ATE for the eight full recordings
with dense references. The dashed line marks unit scale.}
\label{fig:template}
\end{figure}

Figure~\ref{fig:template} also relates baseline scale to aligned ATE. Medium12
has a fitted scale of 0.9508 and ATE 7.01 m; Lowlight10 has scale 0.9997 but
ATE 5.36 m. Near-unit global scale does not imply an accurate trajectory.
Conversely, a nonunit scale alone does not identify the source of drift.

\subsection{Feature density and additional refinement}
\begin{figure}[t]
\centering\includegraphics[width=\linewidth]{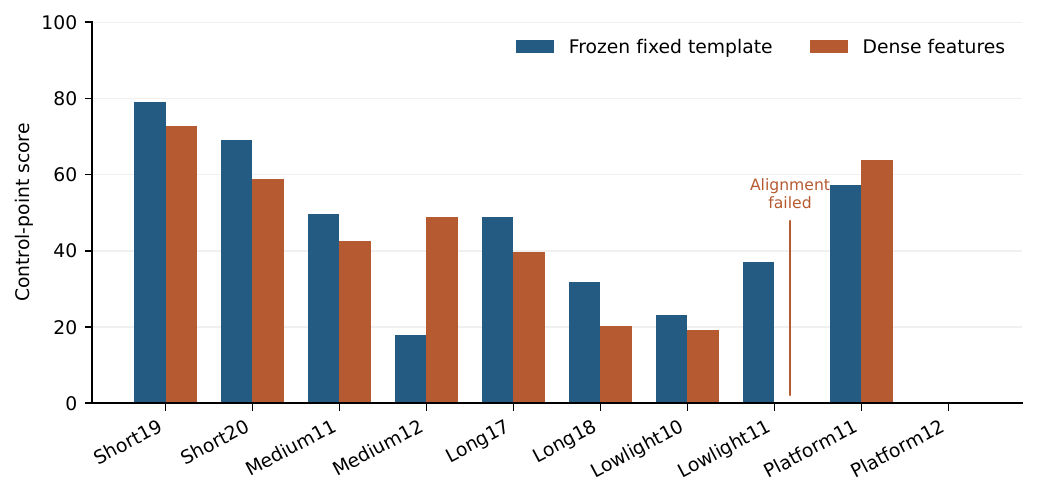}
\caption{Original and dense-feature CP scores on all ten recordings.
The absent Lowlight11 dense bar is an alignment failure, not a measured zero.
Zero scores on Platform12 are actual reported scores.}
\label{fig:density}
\end{figure}

Reducing detection-grid size from 50 to 35 pixels improves Medium12 from 17.83
to 48.98 CP, but reduces Short19 from 79.02 to 72.62, Short20 from 69.08 to
58.81, and Long18 from 31.85 to 20.32. Lowlight11 fails control-point alignment.
Figure~\ref{fig:density} and Table~\ref{tab:density} retain that failure explicitly;
there is no complete-case average that removes it. Both platform variants score
zero on Platform12, even though the dense variant's maximum step is smaller.
Bounding the trajectory is not sufficient for correct localization.

A separate four-record global-reconstruction package yields CP changes of
$+0.79$, $-2.91$, $+2.01$ and $-0.97$ on Short19, Medium12, Long17 and
Lowlight10 relative to the original mono trajectories. This package changes
multiple backend components together; it does not isolate each component's
effect. Its regressions prevent a general improvement claim. Other rotating-world
and recovery studies likewise did not establish an accepted replacement and
are not part of the frozen candidate. Their saved result tables are included
with the paper's evidence data, without being pooled into its baseline scores.

\subsection{Start sensitivity}
\begin{figure}[t]
\centering\includegraphics[width=\linewidth]{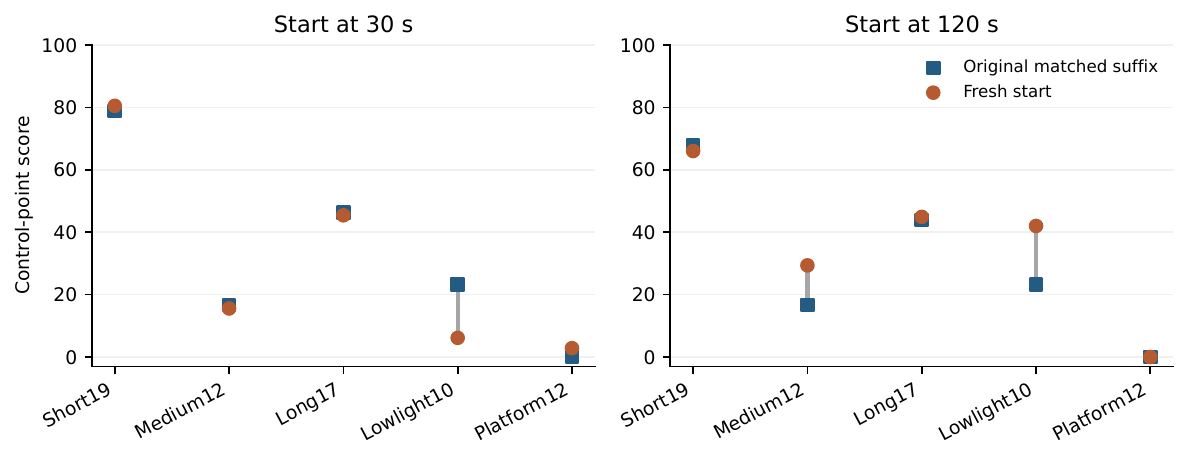}
\caption{CP changes when starting fresh at 30 s and 120 s. Squares represent
the original matched suffix; circles represent fresh runs. The large Lowlight10
change reverses direction across offsets.}
\label{fig:starts}
\end{figure}

\begin{table}[t]
\centering\small\begin{tabular}{lrrrrr}\toprule & & \multicolumn{2}{c}{CP} & \multicolumn{2}{c}{R@5m (\%)} \\ Record & Start (s) & Fresh & Suffix & Fresh & Suffix \\\midrule
Short19 & 30 & 80.49 & 78.82 & 95.00 & 95.00 \\
Medium12 & 30 & 15.55 & 16.72 & 43.72 & 43.26 \\
Long17 & 30 & 45.40 & 46.21 & 90.50 & 90.94 \\
Lowlight10 & 30 & 6.14 & 23.24 & 14.38 & 64.80 \\
Platform12 & 30 & 2.85 & 0.00 & \textemdash & \textemdash \\
Short19 & 120 & 66.00 & 67.83 & 84.83 & 84.83 \\
Medium12 & 120 & 29.36 & 16.72 & 66.97 & 43.26 \\
Long17 & 120 & 44.92 & 43.89 & 87.50 & 87.43 \\
Lowlight10 & 120 & 42.00 & 23.14 & 89.86 & 64.78 \\
Platform12 & 120 & 0.00 & 0.00 & \textemdash & \textemdash \\
\bottomrule
\end{tabular}

\caption{Fresh processes versus the original trajectory restricted to the
same suffix. Both recalls retain the official full-reference denominator.
No start-time winner is selected.}
\label{tab:starts}
\end{table}
All ten controlled starts complete their sensor, coverage and official-rescore
audits. Accuracy nevertheless varies sharply (Table~\ref{tab:starts} and
Figure~\ref{fig:starts}). On Lowlight10, the fresh 30 s start yields 6.14 CP
against 23.24 for the original suffix, with ATE 12.27 versus 5.05 m. At 120 s,
fresh CP rises to 42.00 and ATE falls to 2.00 m. Medium12 similarly changes
direction: fresh CP is below its matched suffix at 30 s but above it at 120 s.

These changes establish sensitivity to the selected input start. They do not
isolate an initialization defect, because changing the start also changes the
subsequent visual history and estimator decisions. Selecting the favorable
offset would conceal this dependence rather than solve it.

Platform12 diverges at both starts. Maximum native steps are 152.54 and 157.62 m
at 30 and 120 s. A CP score of 2.85 at the earlier start is not evidence of
recovery. All 270,564 suffix timestamps remain present, including 201 initial
fills and two later fills. This again separates execution and format integrity
from accurate motion estimation.

\subsection{Runtime}
The ten controlled starts consume 5,230.51 CPU seconds and 3,126.03 wall seconds
in total. Maximum recorded native RSS is 295,492 KiB, about 288.6 MiB.
Per-run native batch throughput ranges from 79.94 to 90.27 images/s
(Table~\ref{tab:runtime}). These runs include observation dumping and exclude
input preparation and official scoring. They show that these shortened native
runs process more than the nominal 20 image timestamps/s under the stated
conditions, but do not establish online latency, end-to-end throughput or
embedded feasibility. The public submission form reports 84.64 frames/s,
computed from the historical full-training total of 265,694 timestamps in
3,139.15 native wall seconds \cite{anchorflow_method}. Those historical runs had
different CPU-affinity conditions and must not be confused with the controlled
starts or a measured main-test processing rate. They also include failed trajectories, so speed must be
interpreted alongside accuracy.

\subsection{Public source release and reproduction checks}
The released code and frozen configurations are publicly available
\cite{anchorflow_source}:
\begin{center}\small
\url{https://github.com/sudhirpratapyadav/anchorflow-vio}
\end{center}
Original AnchorFlow contributions are MIT licensed;
Basalt and other dependencies retain their licences and notices. The build
reconstructs Basalt from pinned upstream commit \texttt{0f3b2b5} plus the frozen
patch and one-camera reader, rather than copying the historical executable.

The release validation record reports a fresh source build of 43 native targets
and a complete Short19 reproduction. All 18,352 exported poses and observation
files match the original run byte for byte on the tested host, including
2,948,425 camera-zero observations. Measured cost is 373.30 CPU seconds,
211.13 wall seconds and 199,668 KiB maximum RSS. The preparer reports the
4.804362 ms terminal IMU gap without changing sensor rows, while rejecting gaps
beyond its permitted interval. These are code and artifact-reproduction checks,
not an independent validation of every accuracy result. A clean network
dependency installation and compiler-independent bitwise reproduction were
not established. The source release's documented Python checks also cover
sensor isolation, trajectory formatting, cropped-run rejection and pinned-input
mutation. Its versioned validation record is included with the paper evidence.
\FloatBarrier

\section{Discussion and Limitations}
\paragraph{Useful comparisons can have opposite signs.}
Medium12 benefits from rolling templates and increased density, whereas Long17
loses accuracy under both changes. This pattern does not support a universal
``more persistent features'' or ``more features'' intervention. The temporal
support diagnostic in the accompanying run records counts raw feature-ID
overlap and occupied image cells; these quantities are not accepted landmark
influence, conditioned depth, or independent inertial observability. We do not
use them as a validated switch between policies.

\paragraph{Coverage must be interpreted with the estimator output.}
The benchmark export is deliberately complete, but 204 full-record poses are
filled and Platform12 remains divergent. The first-pose fill also waits for a
later estimate. A complete file therefore certifies neither causal availability
nor successful visual tracking. The large platform steps are sufficient to
identify failure in these runs; attributing that failure to cabin motion,
outliers or inertial excitation requires additional evidence.

\paragraph{Training analysis and submitted test evaluation have different roles.}
The component studies have only two training records per category, with related recordings and possible
shared device or environment effects. Configuration choices inherit extensive
parent-project tuning. The six expansion records were prospectively declared
for the mono density study, but are not independent held-out data from the wider
development process. The four-record rolling comparison is exploratory, and
the start study covers one recording per category. No random-seed uncertainty
or population-level significance is estimated. The 63-record server evaluation
extends coverage to the benchmark's hidden-test split, but still samples the
same benchmark and device family. It does not establish transfer to another
sensor or environment distribution. Test results are reported for the frozen
submission rather than used as an iterative tuning signal.

\paragraph{System attribution and generality.}
The estimator, square root formulation and sensor models have established
precedents. The evidence supports a measured configuration and its failure
analysis, not algorithmic novelty or universal superiority. The server comparison
provides same-benchmark scores for other submitted methods, but does not isolate
the contribution of our patches against an otherwise identical upstream Basalt
run. Such a matched baseline would strengthen component-level attribution. Other sensors, raw
fisheye input, additional datasets and robot hardware have not been evaluated
for this frozen mono candidate. Dense pseudo ground truth is itself a reference
estimate; control-point and trajectory metrics answer related but different
questions.

\section{Conclusion}
AnchorFlow-VIO uses a strictly single-camera Basalt-derived estimator with a
uniform fixed-template configuration. It obtains mean CP 41.38 across ten
training recordings and server test category scores of 47.2 / 37.1 / 24.8 /
35.1 / 15.6 on all 63 test recordings. Its performance varies
substantially across recordings, and complete pose output coexists with a
divergent moving-platform estimate. Paired comparisons show that fixed templates
and increased feature density have recording-dependent effects. Fresh starts
further reveal large low-light sensitivity. The retained system is a reproducible
open-source system with documented limitations. The submitted test evaluation
extends the training characterization, while its platform failures retain the
need for improvements that survive unfavorable starts and changes in motion
conditions. Broader robustness and cross-sensor transfer remain open.

\section*{Data and Reproducibility}
The accompanying local source package contains the result CSVs, frozen policies,
run receipts, source hashes, full-rate estimated positions, plotting script and
generated figures. Trajectory-derived figures use the verified training archive;
the test figures use recorded server scores. No new SLAM runs were launched for
this manuscript revision. The complete training archive SHA-256 is recorded in
the evidence manifest. Research source snapshot commit is
\texttt{46c98a5}; per-experiment source commits and exact run commands remain in
the supplied experiment and log records. Public source snapshot commit is
\texttt{8ba93ea}. The release pins the upstream estimator, patch and dependency
versions. Original contributions are MIT licensed, with upstream notices retained.
Dataset images and pseudo ground-truth files are not redistributed in this package.

The combined submitted test archive contains all 63 original trajectories and
has SHA-256
\begin{center}\footnotesize
\path{a98232b2f6a7026b6d5d9c5357bb661b162ba0401f54a340d6a5c8e04c0c0754}.
\end{center}
It was checked against the saved completion receipt during this revision.
The public method page and leaderboard provide the external result record
\cite{anchorflow_method,lamaria_leaderboard}. The evidence bundle preserves
the dated category snapshot, all per-sequence displayed scores and their
source provenance. Test reference labels are not available locally.

\bibliographystyle{plainnat}
\bibliography{references}
\FloatBarrier
\appendix
\section{Additional Tables and Reproduction}
\begin{table}[ht]
\centering\small\begin{tabular}{lrrr}\toprule Record & Original CP & Dense CP & Dense $\Delta$ CP \\\midrule
Short19 & 79.02 & 72.62 & -6.40 \\
Short20 & 69.08 & 58.81 & -10.27 \\
Medium11 & 49.53 & 42.46 & -7.07 \\
Medium12 & 17.83 & 48.98 & +31.15 \\
Long17 & 48.90 & 39.81 & -9.09 \\
Long18 & 31.85 & 20.32 & -11.53 \\
Lowlight10 & 23.18 & 19.22 & -3.97 \\
Lowlight11 & 37.13 & \textemdash & Alignment failed \\
Platform11 & 57.30 & 63.78 & +6.48 \\
Platform12 & 0.00 & 0.00 & +0.00 \\
\bottomrule
\end{tabular}

\caption{Complete feature-density comparison. The failed Lowlight11 alignment
is retained as missing, while Platform12 contributes its measured zero.}
\label{tab:density}
\end{table}
\begin{table}[ht]
\centering\small\begin{tabular}{lrrrr}\toprule Record / start & CPU (s) & Wall (s) & RSS (MiB) & Images/s \\\midrule
Short19 / 30 s & 350.87 & 198.35 & 189.5 & 89.50 \\
Medium12 / 30 s & 628.27 & 378.82 & 270.9 & 86.88 \\
Long17 / 30 s & 676.92 & 395.03 & 288.6 & 89.21 \\
Lowlight10 / 30 s & 548.63 & 347.87 & 243.1 & 79.94 \\
Platform12 / 30 s & 498.20 & 294.12 & 246.5 & 88.62 \\
Short19 / 120 s & 317.35 & 179.78 & 180.9 & 88.73 \\
Medium12 / 120 s & 601.69 & 365.78 & 265.9 & 85.06 \\
Long17 / 120 s & 642.67 & 375.02 & 281.5 & 89.17 \\
Lowlight10 / 120 s & 511.45 & 322.46 & 236.2 & 80.66 \\
Platform12 / 120 s & 454.46 & 268.80 & 237.3 & 90.27 \\
\bottomrule
\end{tabular}

\caption{Native controlled-start cost. RSS is the per-run process maximum, not
aggregate service memory. Throughput excludes preparation and scoring.}
\label{tab:runtime}
\end{table}

The supplied \texttt{data/} directory preserves original metric precision.
\path{validation_scores.csv} contains full-record original and dense runs;
\path{frontend_scores.csv} contains the template pairs;
\path{controlled_starts.csv} contains all fresh/suffix comparisons.
Additional CSVs preserve global-refinement, rotating-world and recovery results.
The source protocol and native configuration are copied alongside them.
\path{provenance.json} links each snapshot to its source path and SHA-256.

Figures and tables can be regenerated from these snapshots with
\begin{verbatim}
python3 build_assets.py
latexmk -pdf -interaction=nonstopmode -halt-on-error main.tex
\end{verbatim}
This reproduces the manuscript assets, not fresh estimator measurements.
Fresh mono inference requires calibrated ASL inputs and the pinned native
runtime. In the public source release, the corresponding commands are
\begin{verbatim}
.venv/bin/python scripts/prepare_input.py DATA INPUT
.venv/bin/python scripts/run_sequence.py INPUT RUN \
  --runtime build/native/runtime
\end{verbatim}
Here \texttt{DATA}, \texttt{INPUT} and \texttt{RUN} denote the recording and new
owned output directories. The original bounded process envelope must be applied
for comparable execution conditions. Supplying \texttt{--sequence DATA} adds
training evaluation after trajectory creation. The paper source archive does
not bundle the native runtime or dataset and is not a self-contained estimator
installation. Public build instructions reconstruct the native runtime separately.
\FloatBarrier
\section{Complete Server Test Results}
\label{sec:test_appendix}
\small
\begin{longtable}{lrrr}
\caption{All 63 server-evaluated test recordings. Values reproduce the server display to one decimal; dashes are unavailable metrics. The single reported platform recall of zero is retained as displayed, not generalized to the other platform recordings.}\label{tab:test_sequences}\\
\toprule Sequence & CP & CP R@1m (\%) & Pose R@5m (\%) \\
\midrule\endfirsthead
\multicolumn{4}{l}{\small Table \thetable\ continued}\\
\toprule Sequence & CP & CP R@1m (\%) & Pose R@5m (\%) \\
\midrule\endhead
\midrule\multicolumn{4}{r}{\small Continued on next page}\\\endfoot
\bottomrule\endlastfoot
sequence\_1\_1 & 46.3 & 16.7 & 100.0 \\
sequence\_1\_2 & 22.9 & 0.0 & 55.2 \\
sequence\_1\_3 & 26.5 & 0.0 & 72.6 \\
sequence\_1\_4 & 25.5 & 6.7 & 59.6 \\
sequence\_1\_5 & 60.0 & 50.0 & 99.7 \\
sequence\_1\_6 & 35.4 & 18.2 & 76.2 \\
sequence\_1\_7 & 49.4 & 21.4 & 100.0 \\
sequence\_1\_8 & 70.4 & 85.7 & 100.0 \\
sequence\_1\_9 & 58.6 & 46.2 & 100.0 \\
sequence\_1\_10 & 66.7 & 66.7 & 100.0 \\
sequence\_1\_11 & 75.5 & 100.0 & 100.0 \\
sequence\_1\_12 & 26.1 & 0.0 & 66.0 \\
sequence\_1\_13 & 36.3 & 7.1 & 86.7 \\
sequence\_1\_14 & 68.5 & 66.7 & 100.0 \\
sequence\_1\_15 & 68.6 & 71.4 & 100.0 \\
sequence\_1\_16 & 13.4 & 0.0 & 27.8 \\
sequence\_1\_17 & 44.6 & 26.7 & 79.9 \\
sequence\_1\_18 & 55.4 & 20.0 & 100.0 \\
sequence\_2\_1 & 31.0 & 4.8 & 75.8 \\
sequence\_2\_2 & 25.4 & 9.1 & 63.6 \\
sequence\_2\_3 & 16.2 & 5.9 & 30.9 \\
sequence\_2\_4 & 27.3 & 5.3 & 68.6 \\
sequence\_2\_5 & 30.2 & 5.0 & 58.0 \\
sequence\_2\_6 & 35.5 & 6.3 & 87.4 \\
sequence\_2\_7 & 1.9 & 0.0 & 6.2 \\
sequence\_2\_8 & 73.4 & 75.0 & 100.0 \\
sequence\_2\_9 & 56.1 & 44.4 & 100.0 \\
sequence\_2\_10 & 74.0 & 94.4 & 100.0 \\
sequence\_3\_1 & 21.2 & 3.7 & 34.7 \\
sequence\_3\_2 & 17.2 & 0.0 & 34.3 \\
sequence\_3\_3 & 10.4 & 0.0 & 24.0 \\
sequence\_3\_4 & 6.6 & 0.0 & 15.1 \\
sequence\_3\_5 & 34.5 & 3.7 & 82.1 \\
sequence\_3\_6 & 49.0 & 14.8 & 100.0 \\
sequence\_3\_7 & 14.1 & 0.0 & 23.8 \\
sequence\_3\_8 & 21.8 & 0.0 & 62.0 \\
sequence\_3\_9 & 17.2 & 0.0 & 33.9 \\
sequence\_3\_10 & 21.9 & 0.0 & 46.2 \\
sequence\_3\_11 & 21.6 & 0.0 & 48.1 \\
sequence\_3\_12 & 19.9 & 0.0 & 50.3 \\
sequence\_3\_13 & 22.9 & 0.0 & 43.9 \\
sequence\_3\_14 & 33.8 & 3.8 & 87.0 \\
sequence\_3\_15 & 51.6 & 23.1 & 95.6 \\
sequence\_3\_16 & 32.7 & 3.7 & 82.3 \\
sequence\_4\_1 & 26.2 & 0.0 & 66.8 \\
sequence\_4\_2 & 6.3 & 6.3 & 3.6 \\
sequence\_4\_3 & 6.4 & 0.0 & 13.5 \\
sequence\_4\_4 & 32.5 & 6.3 & 88.4 \\
sequence\_4\_5 & 9.7 & 0.0 & 14.5 \\
sequence\_4\_6 & 60.6 & 53.3 & 100.0 \\
sequence\_4\_7 & 62.8 & 56.3 & 100.0 \\
sequence\_4\_8 & 41.8 & 14.3 & 79.3 \\
sequence\_4\_9 & 69.8 & 92.3 & 100.0 \\
sequence\_5\_1 & 9.2 & 0.0 & \textemdash \\
sequence\_5\_2 & 39.6 & 7.4 & \textemdash \\
sequence\_5\_3 & 12.8 & 3.7 & \textemdash \\
sequence\_5\_4 & 0.0 & 0.0 & 0.0 \\
sequence\_5\_5 & 0.0 & 0.0 & \textemdash \\
sequence\_5\_6 & 8.4 & 0.0 & \textemdash \\
sequence\_5\_7 & 16.5 & 6.7 & \textemdash \\
sequence\_5\_8 & 56.2 & 46.7 & \textemdash \\
sequence\_5\_9 & 2.0 & 0.0 & \textemdash \\
sequence\_5\_10 & 10.9 & 0.0 & \textemdash \\
\end{longtable}

\normalsize
\end{document}